\documentclass[11pt,a4paper]{article}
\usepackage[hyperref]{acl2018}

\usepackage{times}
\usepackage{latexsym}

\usepackage{url}

\usepackage{algorithm}
\usepackage[noend]{algpseudocode}
\usepackage{xcolor,colortbl}
\usepackage{multirow}

\usepackage[utf8]{inputenc} % allow utf-8 input
\usepackage[T1]{fontenc}    % use 8-bit T1 fonts
\usepackage{url}            % simple URL typesetting
\usepackage{booktabs}       % professional-quality tables
\usepackage{amsfonts}       % blackboard math symbols
\usepackage{nicefrac}       % compact symbols for 1/2, etc.
\usepackage{microtype}      % microtypography
\usepackage{amsmath}
\usepackage{graphicx}
\usepackage{subcaption}

\makeatletter
\def\BState{\State\hskip-\ALG@thistlm}
\makeatother

\DeclareMathOperator*{\argmax}{arg\,max}

\aclfinalcopy % Uncomment this line for the final submission
\def\aclpaperid{168} %  Enter the acl Paper ID here

\title{Strong Baselines for Neural Semi-supervised Learning\\under Domain Shift}
\author
{
	\begin{tabular}{cc}
	Sebastian Ruder$^{\spadesuit\clubsuit}$ & Barbara Plank$^{\heartsuit\Diamond}$ \\
	\end{tabular}
	\\
    $^\spadesuit$Insight Research Centre, National University of Ireland, Galway, Ireland\\
    $^\clubsuit$Aylien Ltd., Dublin, Ireland\\
    $^\heartsuit$Center for Language and Cognition, University of Groningen, The Netherlands\\
    $^\Diamond$Department of Computer Science, IT University of Copenhagen, Denmark\\
	{\tt \small{sebastian@ruder.io,bplank@gmail.com}}
}

\date{}

\begin{document}
\maketitle
\begin{abstract}
Novel neural models have been proposed in recent years for learning under domain shift. Most models, however, only evaluate on a single task, on proprietary datasets, or compare to weak baselines, which makes comparison of models difficult. 
In this paper, we re-evaluate classic general-purpose bootstrapping approaches in the context of neural networks under domain shifts vs. recent neural approaches
and propose a novel \emph{multi-task tri-training} method that reduces the time and space complexity of classic tri-training.
Extensive experiments on two benchmarks 
are negative: while our novel method establishes a new state-of-the-art for sentiment analysis, it does not fare consistently the best. More importantly, we arrive at the somewhat surprising conclusion that classic tri-training, with some additions, outperforms the state of the art. We conclude that classic approaches constitute an important and strong baseline.
\end{abstract}

\section{Introduction}

Deep neural networks (DNNs) excel at learning from labeled data and have achieved state of the art in a wide array of supervised NLP tasks such as dependency parsing \cite{Dozat2017}, named entity recognition \cite{Lample2016}, and semantic role labeling \cite{He2017}.

In contrast, learning from unlabeled data, especially under domain shift, remains a challenge. This is common in many real-world applications where the distribution of the training and test data differs. Many state-of-the-art domain adaptation approaches leverage task-specific characteristics such as sentiment words \cite{Blitzer2006,Wu2016a} or distributional features \cite{Schnabel2014,yin-schnabel-schutze:2015:EMNLP} which do not generalize to other tasks. Other approaches that are in theory more general only evaluate on proprietary datasets \cite{Kim2017b} or on a single benchmark \cite{Zhou2016}, which carries the risk of overfitting to the task. In addition, most models only compare against weak baselines and, strikingly, almost none considers evaluating against approaches from the extensive semi-supervised learning (SSL) literature \cite{Chapelle2006}.

In this work, we make the argument that such algorithms make strong baselines for any task in line with recent efforts highlighting the usefulness of classic approaches 
\cite{Melis2017,Denkowski:Neubig:2017}. We re-evaluate bootstrapping algorithms in the context of DNNs. These are general-purpose semi-supervised algorithms that treat the model as a black box and can thus be used easily---with a few additions---with the current generation of NLP models. Many of these methods, though, were originally developed with in-domain performance in mind, so their effectiveness in a domain adaptation setting remains unexplored.

In particular, we re-evaluate three traditional bootstrapping methods, self-training \cite{Yarowsky1995}, tri-training \cite{Zhou2005}, and tri-training with disagreement \cite{Sogaard2010} for neural network-based approaches on \textit{two} NLP tasks with different characteristics, namely, a sequence prediction and a classification task (POS tagging and sentiment analysis). We evaluate the methods across multiple domains on two well-established benchmarks, without taking any further task-specific measures, and compare to the best results published in the literature.

We make the somewhat surprising observation that classic tri-training outperforms task-agnostic state-of-the-art semi-supervised learning \cite{Laine2017} and recent neural adaptation approaches~\cite{Ganin2016,Saito2017}.

In addition, we propose \emph{multi-task tri-training}, which reduces the main deficiency of tri-training, namely its time and space complexity. It establishes a new state of the art on unsupervised domain adaptation for sentiment analysis but it is outperformed by classic tri-training for POS tagging.

\paragraph{Contributions} Our contributions are: a) We propose a novel multi-task tri-training method. b) We show that tri-training can serve as a strong and robust semi-supervised learning baseline for the current generation of NLP models.\ c) We perform an extensive evaluation of bootstrapping\footnote{We use the term bootstrapping as used in the semi-supervised learning literature \cite{Zhu2005}, which should not be confused with the statistical procedure of the same name \cite{efron1994introduction}.} algorithms compared to state-of-the-art approaches on two benchmark datasets. d) We shed light on the task and data characteristics that yield the best performance for each model.

\section{Neural bootstrapping methods} \label{sec:neural_boostrapping}

We first introduce three classic bootstrapping methods, self-training, tri-training, and tri-training with disagreement and detail how they can be used with neural networks. For in-depth details we refer the reader to~\cite{Abney2007,Chapelle2006,Zhu:Goldberg:2009}. We introduce our novel multi-task tri-training method in \S\ref{sec:mt-tri}.

\subsection{Self-training}

Self-training \cite{Yarowsky1995,McClosky2006} is one of the earliest and simplest bootstrapping approaches. In essence, it leverages the model's own predictions on unlabeled data to obtain additional information that can be used during training. Typically the most confident predictions are taken at face value, as detailed next.

Self-training trains a model $m$ on a labeled training set $L$ and an unlabeled data set $U$. At each iteration, the model provides predictions $m(x)$ in the form of a probability distribution over classes for all unlabeled examples $x$ in $U$. If the probability assigned to the most likely class is higher than a predetermined threshold $\tau$, $x$ is added to the labeled examples with $p(x) = \argmax m(x)$ as pseudo-label. This instantiation is the most widely used and shown in Algorithm \ref{alg:self-training}. 

\begin{algorithm}\centering
\caption{Self-training \cite{Abney2007}} \label{alg:self-training}
\begin{algorithmic}[1]
%\Procedure{TriTrain}{}
\Repeat
	\State $m \gets train\_model(L)$
	\For {$x \in U$}
    	\If {$\max m(x) > \tau$}
        	\State $L \gets L \cup \{(x, p(x))\}$
    	\EndIf
    \EndFor
\Until {no more predictions are confident}
%\EndProcedure
\end{algorithmic}
\end{algorithm}

\paragraph{Calibration} It is well-known that output probabilities in neural networks are poorly calibrated \cite{Guo2017}. Using a fixed threshold $\tau$ is thus not the best choice. While the \emph{absolute} confidence value is inaccurate, we can expect that the \emph{relative} order of confidences is more robust.

For this reason, we select the top $n$ unlabeled examples that have been predicted with the highest confidence after every epoch and add them to the labeled data. This is one of the many variants for self-training, called \textit{throttling}~\cite{Abney2007}.
We empirically confirm that this outperforms the classic selection in our experiments.

\paragraph{Online learning} In contrast to many classic algorithms, DNNs are trained online by default. We compare training setups and find that training until convergence on labeled data and then training until convergence using self-training performs best.

Classic self-training has shown mixed success. In parsing it proved successful only with small datasets~\cite{reichart2007self} or when a generative component is used together with a reranker in high-data conditions~\cite{McClosky2006,suzuki2008semi}. Some success was achieved with careful task-specific data selection~\cite{Petrov2012}, while others report limited success on a variety of NLP tasks~\cite{
Plank2011,VanAsch2016,Vandergoot2017}. 
Its main downside is that the model is not able to correct its own mistakes and errors are amplified, an effect that is increased under domain shift.

\subsection{Tri-training}

Tri-training \cite{Zhou2005} is a classic method that reduces the bias of predictions on unlabeled data by utilizing the agreement of three independently trained models. Tri-training (cf.\ Algorithm \ref{alg:tri-training}) first trains three models $m_1$, $m_2$, and $m_3$ on bootstrap samples of the labeled data $L$. An unlabeled data point is added to the training set of a model $m_i$ if the other two models $m_j$ and $m_k$ agree on its label. Training stops when the classifiers do not change anymore. 

\begin{algorithm}
\caption{Tri-training \cite{Zhou2005}}\label{alg:tri-training}
\begin{algorithmic}[1]
%\Procedure{TriTrain}{}
\For {$i \in \{1..3\}$}
\State $S_i \gets bootstrap\_sample(L)$
\State $m_i \gets train\_model(S_i)$
\EndFor
\Repeat
	\For {$i \in \{1..3\}$}
        \State $L_i \gets \emptyset$
		\For {$x \in U$}
            \If {$p_j(x) = p_k(x)(j,k \neq i)$}
            	\State $L_i \gets L_i \cup \{(x, p_j(x))\}$
            \EndIf
        \EndFor
        $m_i \gets train\_model(L \cup L_i)$
	\EndFor
\Until {none of $m_i$ changes}
\State apply majority vote over $m_i$
%\EndProcedure
\end{algorithmic}
\end{algorithm}

Tri-training \textit{with disagreement} \cite{Sogaard2010} is based on the intuition that a model should only be strengthened in its weak points and that the labeled data should not be skewed by easy data points. In order to achieve this, it adds a simple modification to the original algorithm (altering line 8 in Algorithm \ref{alg:tri-training}), requiring that for an unlabeled data point on which $m_j$ and $m_k$ \emph{agree}, the other model $m_i$ \emph{disagrees} on the prediction. Tri-training with disagreement is more data-efficient than tri-training and has achieved competitive results on part-of-speech tagging \cite{Sogaard2010}.

\paragraph{Sampling unlabeled data} Both tri-training and tri-training with disagreement can be very expensive in their original formulation as they require to produce predictions for each of the three models on all unlabeled data samples, which can be in the millions in realistic applications. We thus propose to sample a number of unlabeled examples at every epoch. For all traditional bootstrapping approaches we sample 10k candidate instances in each epoch. For the neural approaches we use a linearly growing candidate sampling scheme proposed by~\cite{Saito2017},  increasing the candidate pool size as the models become more accurate.

\paragraph{Confidence thresholding} Similar to self-training, we can introduce an additional requirement that pseudo-labeled examples are only added if the probability of the prediction of at least one model is higher than some threshold $\tau$. We did not find this to outperform prediction without threshold for traditional tri-training, but thresholding proved essential for our method (\S \ref{sec:mt-tri}).

The most important condition for tri-training and tri-training with disagreement is that the models are diverse. Typically, bootstrap samples are used to create this diversity \cite{Zhou2005,Sogaard2010}. However, training separate models on bootstrap samples of a potentially large amount of training data is expensive and takes a lot of time. This drawback motivates our approach.

\subsection{Multi-task tri-training} \label{sec:mt-tri}

In order to reduce both the time and space complexity of tri-training, we propose Multi-task Tri-training (MT-Tri). MT-Tri leverages insights from multi-task learning (MTL) \cite{Caruana1993} to share knowledge across models and accelerate training. Rather than storing and training each model separately, we propose to share the parameters of the models and train them jointly using MTL.\footnote{Note: we use the term multi-task learning here albeit all tasks are of the same kind, similar to work on multi-lingual modeling treating each language (but same label space) as separate task e.g.,~\cite{fang2017model}. It is interesting to point out that our model is  further doing implicit multi-view learning by way of the orthogonality constraint.} All models thus collaborate on learning a joint representation, which improves convergence. 

\begin{figure}[]\centering
\includegraphics[width=0.9\columnwidth]{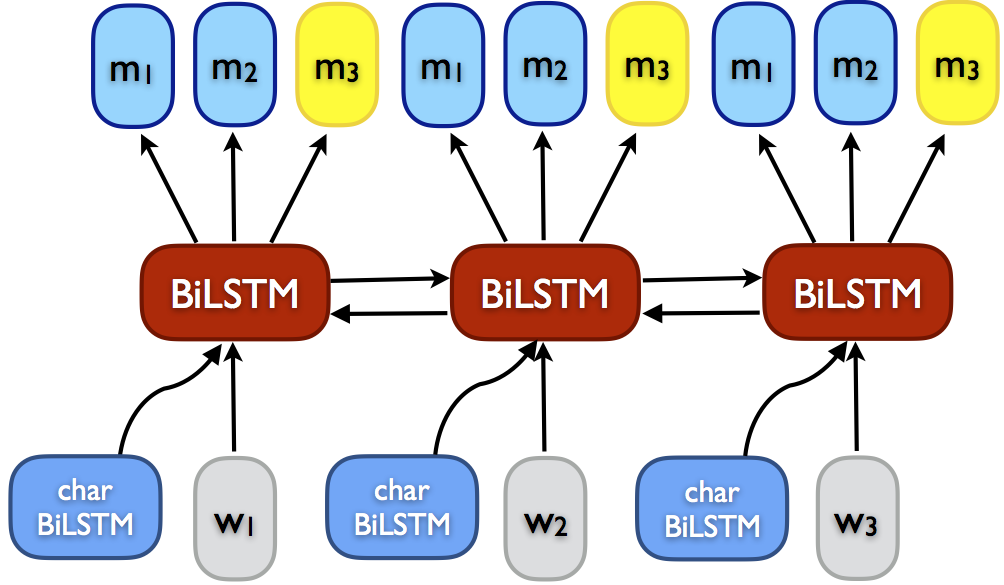}
\caption{Multi-task tri-training (MT-Tri).}
\label{fig:model}
\end{figure}

The output softmax layers are model-specific and are only updated for the input of the respective model. We show the model in Figure \ref{fig:model} (as instantiated for POS tagging). As the models leverage a joint representation, we need to ensure that the features used for prediction in the softmax layers of the different models are as diverse as possible, so that the models can still learn from each other's predictions. In contrast, if the parameters in all output softmax layers were the same, the method would degenerate to self-training.

To guarantee diversity, we introduce an orthogonality constraint \cite{Bousmalis2016} as an additional loss term, which we define as follows:

\begin{equation}
{\cal L}_{orth} = \|W_{m_1}^\top W_{m_2}\|^2_F 
\end{equation}

where $|\ \cdot \|^2_F$ is the squared Frobenius norm and $W_{m_1}$ and $W_{m_2}$ are the softmax output parameters of the two source and pseudo-labeled output layers $m_1$ and $m_2$, respectively. The orthogonality constraint encourages the models not to rely on the same features for prediction. As enforcing pair-wise orthogonality between three matrices is not possible, we only enforce orthogonality between the softmax output layers of $m_1$ and $m_2$,\footnote{We also tried enforcing orthogonality on a hidden layer rather than the output layer, but this did not help.} while $m_3$ is gradually trained to be more target-specific. We parameterize ${\cal L}_{orth}$ by $\gamma$=$0.01$ following~\cite{Liu:ea:2017:ACL}. We do not further tune $\gamma$.

More formally, let us illustrate the model by taking the sequence prediction task (Figure~\ref{fig:model}) as illustration. Given an utterance with labels $y_1,..,y_n$, our Multi-task Tri-training loss consists of three task-specific ($m_1,m_2,m_3$) tagging loss functions (where $\vec{h}$ is the uppermost Bi-LSTM encoding): % in case of POS): 
\begin{align}
{\cal L}({\boldsymbol \theta}) = - \sum_i \sum_{1,..,n} \log P_{m_i}(y|\vec{h}) + \gamma {\cal L}_{orth}
\end{align}

In contrast to classic tri-training, we can train the multi-task model with its three model-specific outputs jointly and \textit{without} bootstrap sampling on the labeled source domain data until convergence, as the orthogonality constraint enforces different representations between models $m_1$ and $m_2$. From this point, we can leverage the pair-wise agreement of two output layers to add pseudo-labeled examples as training data to the third model. We train the third output layer $m_3$ only on pseudo-labeled target instances in order to make tri-training more robust to a domain shift. For the final prediction, majority voting of all three output layers is used, which resulted in the best instantiation, together with confidence thresholding  ($\tau = 0.9$, except for high-resource POS where $\tau =0.8$ performed slightly better). We also experimented with using a domain-adversarial loss \cite{Ganin2016} on the jointly learned representation, but found this not to help. The full pseudo-code is given in Algorithm \ref{alg:mt-tri-training}. 

\begin{algorithm}[t!]
\caption{Multi-task Tri-training}\label{alg:mt-tri-training}
\begin{algorithmic}[1]
\State $m \gets train\_model(L)$
\Repeat
	\For {$i \in \{1..3\}$}
        \State $L_i \gets \emptyset$
		\For {$x \in U$}
            \If {$p_j(x) = p_k(x)(j,k \neq i)$}
            	\State $L_i \gets L_i \cup \{(x, p_j(x))\}$
            \EndIf
        \EndFor
        \If {$i = 3$}
        	$m_i = train\_model(L_i)$
        \Else
        	{$m_i \gets train\_model(L \cup L_i)$}
        \EndIf
        
	\EndFor
\Until {end condition is met}
\State apply majority vote over $m_i$
%\EndProcedure
\end{algorithmic}
\end{algorithm}

\paragraph{Computational complexity} The motivation for MT-Tri was to reduce the space and time complexity of tri-training. We thus give an estimate of its efficiency gains. MT-Tri is \textasciitilde$3\times$ more space-efficient than regular tri-training; tri-training stores one set of parameters for each of the three models, while MT-Tri only stores one set of parameters (we use three output layers, but these make up a comparatively small part of the total parameter budget). In terms of time efficiency, tri-training first requires to train each of the models from scratch. The actual tri-training takes about the same time as training from scratch and requires a separate forward pass for each model, effectively training three independent models simultaneously. In contrast, MT-Tri only necessitates one forward pass as well as the evaluation of the two additional output layers (which takes a negligible amount of time) and requires about as many epochs as tri-training until convergence (see Table \ref{tab:pos-low-data-results}, second column) while adding fewer unlabeled examples per epoch (see Section \ref{sec:results}). In our experiments, MT-Tri trained about $5$-$6\times$ faster than traditional tri-training.

MT-Tri can be seen as a self-ensembling technique, where different variations of a model are used to create a stronger ensemble prediction. Recent approaches in this line are \textit{snapshot ensembling} \cite{Huang2017c} that ensembles models converged to different minima during a training run, \textit{asymmetric tri-training}~\cite{Saito2017} (\textsc{Asym}) that leverages agreement on two models as information for the third, and \textit{temporal ensembling} \cite{Laine2017}, which ensembles predictions of a model at different epochs. We tried to compare to temporal ensembling in our experiments, but were not able to obtain consistent results.\footnote{We suspect that the sparse features in NLP and the domain shift might be detrimental to its unsupervised consistency loss.} We compare to the closest most recent method, asymmetric tri-training~\cite{Saito2017}. It differs from ours in two aspects: a) \textsc{Asym} leverages only pseudo-labels from data points on which $m_1$ and $m_2$ agree, and b) it uses only one task ($m_3$) as final predictor. In essence, our formulation of MT-Tri is closer to the original tri-training formulation (agreements on two provide pseudo-labels to the third) thereby incorporating more diversity.

\section{Experiments}

\begin{table}[]\centering
\begin{small}
\begin{tabular}{l l rr}
\toprule
 & \textbf{Domain} & \textbf{\# labeled} & \textbf{\# unlabeled} \\
\midrule
\multirow{6}{*}{\rotatebox[origin=c]{90}{POS tagging}} & Answers & 3,489 & 27,274\\
 & Emails & 4,900 & 1,194,173 \\
& Newsgroups & 2,391 & 1,000,000 \\
& Reviews & 3,813 & 1,965,350 \\
& Weblogs & 2,031 & 524,834 \\
& WSJ & 30,060 & 100,000\\ % fix!
\midrule
\multirow{4}{*}{\rotatebox[origin=c]{90}{Sentiment}} & Book & 2,000 & 4,465\\
& DVD & 2,000 & 3,586 \\
& Electronics & 2,000 & 5,681\\
& Kitchen & 2,000 & 5,945\\
\bottomrule
\end{tabular}
\end{small}
\caption{Number of labeled and unlabeled sentences for each domain in the SANCL 2012 dataset \cite{Petrov2012} for POS tagging (above) and the Amazon Reviews dataset \cite{Blitzer2006} for sentiment analysis (below).}
\label{tab:data-stats}
\end{table}

\label{sec:data}
In order to ascertain which methods  are robust across different domains, we evaluate on two widely used unsupervised domain adaptation datasets for two tasks, a sequence labeling and a classification task, cf.\  
Table \ref{tab:data-stats} for data statistics.

\subsection{POS tagging} For POS tagging we use the SANCL 2012 shared task dataset~\cite{Petrov2012} and compare to the top results in both low and high-data conditions~\cite{Schnabel2014,yin-schnabel-schutze:2015:EMNLP}. Both are strong baselines, as the FLORS tagger has been developed for this challenging dataset and it is based on contextual distributional features (excluding the word's identity), and hand-crafted suffix and shape features (including some language-specific morphological features).
We want to gauge to what extent we can adopt a nowadays fairly standard (but more lexicalized) general neural tagger. 

Our POS tagging model is a state-of-the-art Bi-LSTM tagger \cite{Plank2016} with word and 100-dim character embeddings. Word embeddings are initialized with the 100-dim Glove embeddings~\cite{pennington2014glove}. The BiLSTM has one hidden layer with 100 dimensions. The base POS model is trained on WSJ with early stopping on the WSJ development set, using patience 2, Gaussian noise with $\sigma=0.2$ and word dropout with $p=0.25$~\cite{TACL885}. %The model is comparable to alternative taggers on WSJ, see Table~\ref{tbl:POS-dev-test}.

Regarding data, the source domain is the Ontonotes 4.0 release of the Penn treebank Wall Street Journal (WSJ) annotated for 48 fine-grained POS tags. This amounts to 30,060 labeled sentences. We use 100,000 WSJ sentences from 1988 as unlabeled data, following~\newcite{Schnabel2014}.\footnote{Note that our unlabeled data might slightly differ from theirs. We took the first 100k sentences from the 1988 WSJ dataset from the BLLIP 1987-89 WSJ Corpus Release 1.}  As target data, we use the five SANCL domains (answers, emails, newsgroups, reviews, weblogs). We restrict the amount of unlabeled data for each SANCL domain to the first 100k sentences, and 
do not do any pre-processing. We consider the development set of \textsc{Answers} as our only target dev set to set hyperparameters. This may result in suboptimal per-domain settings but better resembles an unsupervised adaptation scenario. 

\subsection{Sentiment analysis} For sentiment analysis, we evaluate on the Amazon reviews dataset \cite{Blitzer2006}. Reviews with 1 to 3 stars are ranked as negative, while reviews with 4 or 5 stars are ranked as positive. The dataset consists of four domains, yielding 12 adaptation scenarios. We use the same pre-processing and architecture as used in \cite{Ganin2016,Saito2017}: 5,000-dimensional tf-idf weighted unigram and bigram features as input; 2k labeled source samples and 2k unlabeled target samples for training, 200 labeled target samples for validation, and between 3k-6k samples for testing. The model is an MLP with one hidden layer with $50$ dimensions, sigmoid activations, and a softmax output. We compare against the
Variational Fair Autoencoder (VFAE) \cite{louizos2015variational} model and domain-adversarial neural networks (DANN) \cite{Ganin2016}.

\begin{figure*}[!hbt]
    % \caption{Global caption}
    \begin{minipage}{.48\linewidth}
      \centering
         \includegraphics[height=1.5in]{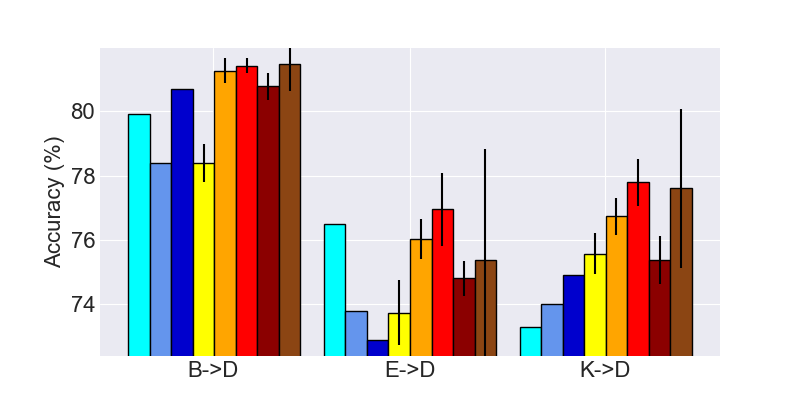}
    \end{minipage}%
    \hspace*{0.1cm}
    \begin{minipage}{.48\linewidth}
      \centering
         \includegraphics[height=1.35in]{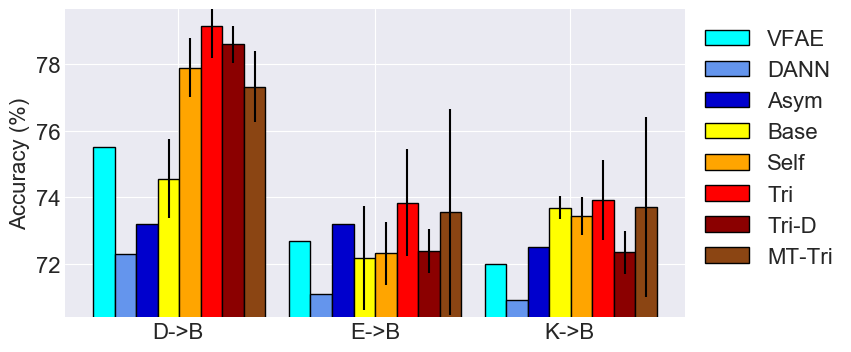}
    \end{minipage}
    % \caption{Global caption}
    \begin{minipage}{.48\linewidth}
      \centering
        \includegraphics[height=1.5in]{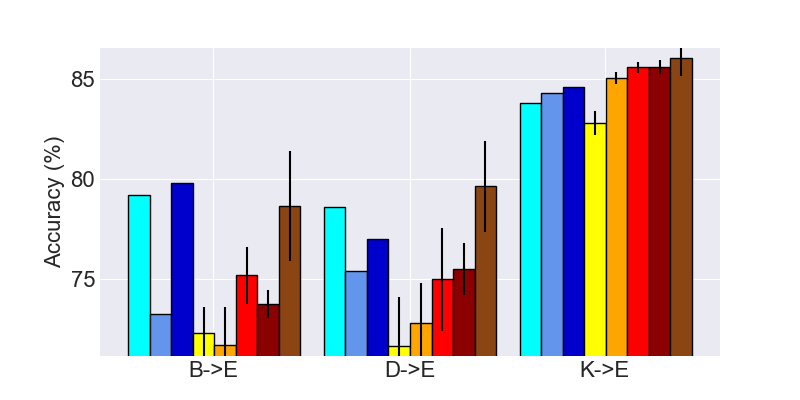}
    \end{minipage}%
    \hspace*{0.2cm}
    \begin{minipage}{.48\linewidth}
      \centering
        \includegraphics[height=1.5in]{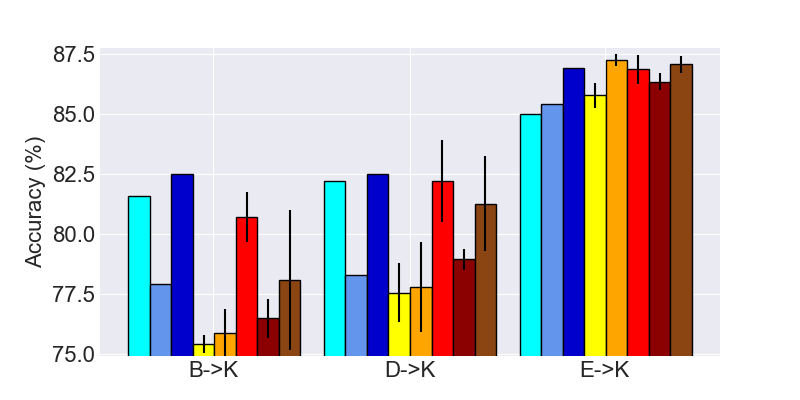}
    \end{minipage}
    \caption{Average results for unsupervised domain adaptation on the Amazon dataset. Domains: B (Book), D (DVD), E (Electronics), K (Kitchen). Results for VFAE, DANN, and Asym are from \newcite{Saito2017}.}
\label{fig:sentiment_results}
\end{figure*}

\subsection{Baselines} 
Besides comparing to the top results published on both datasets, we include the following baselines:
\begin{itemize}
\itemsep-3pt
\item[a)] the task model trained on the source domain; 
\item[b)] self-training (Self); 
\item[c)] tri-training (Tri); 
\item[d)] tri-training with disagreement (Tri-D); and 
\item[e)] asymmetric tri-training \cite{Saito2017}. 
\end{itemize}
Our proposed model is multi-task tri-training (MT-Tri). We implement our models in DyNet \cite{neubig2017dynet}. Reporting single evaluation scores might result in biased results~\cite{reimers-gurevych:2017:EMNLP2017}. Throughout the paper, we report mean accuracy and standard deviation over five runs for POS tagging and over ten runs for sentiment analysis. Significance is computed using bootstrap test. The code for all experiments is released at: \url{https://github.com/bplank/semi-supervised-baselines}.

\subsection{Results} \label{sec:results}

\paragraph{Sentiment analysis} We show results for sentiment analysis for all 12 domain adaptation scenarios in Figure \ref{fig:sentiment_results}. For clarity, we also show the accuracy scores averaged across each target domain as well as a global macro average in Table \ref{tab:sentiment_results}.
\begin{table}[ht!]
\centering
\resizebox{\columnwidth}{!}{%
\begin{tabular}{l c c c c c}
\toprule
\textbf{Model} & \textbf{D} & \textbf{B} & \textbf{E} & \textbf{K} & \textbf{Avg}\\
\midrule
VFAE* & 76.57 & 73.40 & 80.53 & 82.93 & 78.36\\
DANN* & 75.40 & 71.43 & 77.67 & 80.53 & 76.26\\
Asym* & 76.17 & 72.97 & 80.47 & \textbf{83.97} & 78.39\\
\midrule
Src & 75.91 & 73.47 & 75.61 & 79.58 & 76.14\\
Self & 78.00 & 74.55 & 76.54 & 80.30 & 77.35\\
Tri & \textbf{78.72} & \textbf{75.64} & 78.60 & 83.26& 79.05 \\
Tri-D & 76.99 & 74.44 & 78.30 & 80.59 & 77.58\\
MT-Tri & 78.14 & 74.86 & \textbf{81.45} & 82.14 & \textbf{79.15}\\
\bottomrule
\end{tabular}
}
\caption{Average accuracy scores for each SA target domain. *: result from \newcite{Saito2017}.}
\label{tab:sentiment_results}
\end{table}

Self-training achieves surprisingly good results but is not able to compete with tri-training. Tri-training with disagreement is only slightly better than self-training, showing that the disagreement component might not be useful when there is a strong domain shift.
Tri-training achieves the best average results on two target domains and clearly outperforms the state of the art on average.

MT-Tri finally outperforms the state of the art on 3/4 domains, and even slightly traditional tri-training, resulting in the overall best method. This improvement is mainly due to the B->E and D->E scenarios, on which tri-training struggles. These domain pairs are among those with the highest $\mathcal{A}$-distance \cite{Blitzer2007}, which highlights that tri-training has difficulty dealing with a strong shift in domain. Our method is able to mitigate this deficiency by training one of the three output layers only on pseudo-labeled target domain examples.

In addition, MT-Tri is more efficient as it adds a smaller number of pseudo-labeled examples than tri-training at every epoch. For sentiment analysis, tri-training adds around 1800-1950/2000 unlabeled examples at every epoch, while MT-Tri only adds around 100-300 in early epochs. This shows that the orthogonality constraint is useful for inducing diversity. In addition, adding fewer examples poses a smaller risk of swamping the learned representations with useless signals and is more akin to fine-tuning, the standard method for supervised domain adaptation \cite{Howard2018}.
%%%%%% POS 10%

% epoch always set on answers
\begin{table*}[ht!]
\centering
\resizebox{\textwidth}{!}{%
\begin{tabular}{l r c c c c c c c r}
\toprule
& &\multicolumn{5}{c}{\cellcolor[gray]{.85}\textbf{Target domains}} & \\
\textbf{Model}  & $ep$ & \textbf{Answers} & \textbf{Emails} & \textbf{Newsgroups} & \textbf{Reviews} & \textbf{Weblogs} & \textbf{Avg} & \textbf{WSJ} & \textbf{$\mu_{pseudo}$}\\
\midrule
Src (+glove) & & 87.63 $\pm$.37 & 86.49 $\pm$.35 & \textbf{88.60} $\pm$.22 & 90.12 $\pm$.32 & 92.85 $\pm$.17 & 89.14 $\pm$.28 &  95.49 $\pm$.09 & ---\\
Self &(5) & 87.64  $\pm$.18 & 86.58  $\pm$.30 & 88.42  $\pm$.24 & 90.03  $\pm$.11 & 92.80  $\pm$.19 & 89.09 $\pm$.20 & 95.36  $\pm$.07 & .5k\\
Tri &(4) & 88.42 $\pm$.16 & 87.46 $\pm$.20 &  87.97 $\pm$.09 & 90.72 $\pm$.14 & 93.40 $\pm$.15   & 89.56 $\pm$.16 & 95.94 $\pm$.07 & 20.5k \\
Tri-D &(7) & \textbf{88.50} $\pm$.04 & \textbf{87.63} $\pm$.15 & 88.12 $\pm$.05 & \textbf{90.76} $\pm$.10 & \textbf{93.51} $\pm$.06 & \textbf{89.70} $\pm$.08 &  \textbf{95.99} $\pm$.03 & 7.7K\\
Asym & (3) & 87.81 $\pm$.19 & 86.97 $\pm$.17 &  87.74 $\pm$.24 & 90.16 $\pm$.17 & 
92.73 $\pm$.16  & 89.08 $\pm$.19
 & 95.55 $\pm$.12 & 1.5k \\ % epoch 3
MT-Tri & (4) & 87.92 $\pm$.18 & 87.20 $\pm$.23 & 87.73 $\pm$.37 & 90.27 $\pm$.10
 & 92.96 $\pm$.07 & 89.21 $\pm$.19
  & 95.50 $\pm$.06 & 7.6k\\ % epoch 4
\midrule
FLORS & & 89.71 & 88.46 & 89.82 & 92.10 & 94.20 &  90.86 &  95.80 & ---\\
\bottomrule
\end{tabular}%
} % end resizebox
\caption{Accuracy scores on dev set of target domain for POS tagging for 10\% labeled data. Avg: average over the 5 SANCL domains. Hyperparameter $ep$ (epochs) is tuned on Answers dev. $\mu_{pseudo}$: average amount of added pseudo-labeled data. FLORS: results for Batch (u:big) from \cite{yin-schnabel-schutze:2015:EMNLP} (see \S \ref{sec:data}).}
\label{tab:pos-low-data-results}
\end{table*}

%%% POS dev full
\begin{table*}[ht!]
\centering
\resizebox{\textwidth}{!}{%
\begin{tabular}{l c c c c c c c c}
\toprule
 & \multicolumn{5}{c}{\cellcolor[gray]{.85}\textbf{Target domains dev sets}} & \textbf{Avg on} \\
\textbf{Model} & \textbf{Answers} & \textbf{Emails} & \textbf{Newsgroups} & \textbf{Reviews} & \textbf{Weblogs} & \textbf{targets} & \textbf{WSJ} \\
TnT* & 88.55 & 88.14 & 88.66 & 90.40 & 93.33 & 89.82 &95.75\\
Stanford* & 88.92 & 88.68 & 89.11 & 91.43 & 94.15 & 90.46& 96.83\\
%SVMTool* & 88.96 & 88.64 & 89.14 & 91.30 & 94.21 &90.45& 96.63\\
%C\&P* & 89.08 & 88.74 & 89.51 & 91.58 & 94.41 &90.66& 96.78\\
\midrule
Src  & 88.84 $\pm$.15 & 88.24 $\pm$.12 & 89.45 $\pm$.23 & 91.24 $\pm$.03 & 93.92 $\pm$.17 & 90.34 $\pm$.14 & 96.69 $\pm$.08\\
% Self & 88.77  $\pm$.05 & 88.21  $\pm$.11 & 89.32  $\pm$.20 & 91.31  $\pm$.11 & 94.05  $\pm$.08 & 96.58  $\pm$.06\\
Tri & 89.34 $\pm$.18 & 88.83  $\pm$.07 & 89.32  $\pm$.21 & 91.62  $\pm$.06 &  94.40  $\pm$.06 & 90.70 $\pm$.12 & 96.84 $\pm$.04\\  %iter 4 
Tri-D & 89.35 $\pm$.16 &88.66 $\pm$.09 & 89.29 $\pm$.12 & 91.58 $\pm$.05& 94.32 $\pm$.05 & 90.62 $\pm$.09 & 96.85 $\pm$.06 \\ %iter 4
\midrule
Src (+glove) & 89.35 $\pm$.16 & 88.55 $\pm$.14 &  \textbf{90.12} $\pm$.31 & 91.48 $\pm$.15 & 94.48 $\pm$.07 & 90.80 $\pm$.17 & 96.90 $\pm$.04\\
% Self & 89.44 $\pm$.17 & 88.48 $\pm$.24 & 89.87 $\pm$.14 & 91.64 $\pm$.16 & 94.40 $\pm$.14 & 96.72 $\pm$.04\\
Tri  & \textbf{90.00} $\pm$.03 & \textbf{89.06} $\pm$.16 & 90.04 $\pm$.25 & \textbf{91.98} $\pm$.11 & \textbf{94.74} $\pm$.06 & \textbf{91.16} $\pm$.12 & 	\textbf{96.99} $\pm$.02\\
Tri-D &  89.80 $\pm$.19 & 88.85 $\pm$.10 &  90.03 $\pm$.22 & \textbf{91.98} $\pm$.09 & 94.70 $\pm$.05 & 91.01 $\pm$.13 & 96.95 $\pm$.05
\\
Asym   & 89.51 $\pm$.15 & 88.47 $\pm$.19 & 89.26 $\pm$.16 & 91.60 $\pm$.20 & 94.28 $\pm$.15 & 90.62 $\pm$.17 & 96.56 $\pm$.01\\ % epoch 3
% MT-Tri & 89.38 $\pm$.22 & 88.48 $\pm$.17 & 89.63 $\pm$.14 & 91.49 $\pm$.35 & 94.30 $\pm$.17 & 90.66 $\pm$.21\\ = c.09
MT-Tri  & 89.45 $\pm$.05 & 88.65 $\pm$.04 & 89.40 $\pm$.22& 91.63 $\pm$.23 & 94.41 $\pm$.05 & 90.71 $\pm$.12 & 97.37 $\pm$.07\\ % = c.08
	
\midrule 
FLORS*  & 90.30 & 89.44 & 90.86 & 92.95 & 94.71 & 91.66 & 96.59\\
\bottomrule
& \multicolumn{5}{c}{\cellcolor[gray]{.85}\textbf{Target domains test sets}} & \textbf{Avg on} \\
\textbf{Model} & \textbf{Answers} & \textbf{Emails} & \textbf{Newsgroups} & \textbf{Reviews} & \textbf{Weblogs} & \textbf{targets} & \textbf{WSJ} \\
TnT* & 89.36 & 87.38 & 90.85 & 89.67 & 91.37 & 89.73 & 96.57 \\
Stanford* & 89.74 & 87.77 & 91.25 & 90.30 & 92.32 & 90.28 & 97.43 \\
%SVMTool* & 89.90 & 87.74 & 91.21 & 90.01 & 92.05 & 90.18& 97.26\\
%C\&P* & 89.90 & 87.91 & 91.68 & 90.42 & 92.22 & 90.42 & 97.44\\
\midrule
Src (+glove) & 90.43 $\pm$.13 & 87.95 $\pm$.18 & 91.83 $\pm$.20 & 90.04 $\pm$.11& 92.44 $\pm$.14 & 90.54 $\pm$.15 & \textbf{97.50} $\pm$.03\\ 
%Self & 90.44 $\pm$.13 & 87.88 $\pm$.28 & 91.51 $\pm$.18 & 89.94 $\pm$.22 & 92.35 $\pm$.09 & 97.32 $\pm$.07\\
Tri & \textbf{91.21} $\pm$.06 & \textbf{88.30} $\pm$.19 & \textbf{92.18} $\pm$.19 & \textbf{90.06} $\pm$.10 & \textbf{92.85} $\pm$.02 & \textbf{90.92} $\pm$.11 &  97.45 $\pm$.03
\\ %+glove
% Tri-D & 91.02 $\pm$.23 & 88.09 $\pm$.06 & 92.04 $\pm$.12 & 90.21 $\pm$.13 & 92.87 $\pm$.01\\
Asym   & 90.62 $\pm$.26 & 87.71 $\pm$.07 & 91.40 $\pm$.05 & 89.89 $\pm$.22 & 92.37 $\pm$.27 & 90.39 $\pm$.17 & 97.19 $\pm$.03\\
MT-Tri & 90.53 $\pm$.15 & 87.90  $\pm$.07 & 91.45 $\pm$.19 & 89.77 $\pm$.26 & 92.35 $\pm$.09 & 90.40 $\pm$.15 & 97.37 $\pm$.07 \\
\midrule
FLORS* & 91.17 & 88.67 & 92.41 & 92.25 & 93.14 & 91.53 & 97.11\\
\bottomrule
\end{tabular}%
} % end resizebox
\caption{Accuracy for POS tagging  on the dev and test sets of the SANCL domains, models trained on full source data setup. Values for methods with * are from \cite{Schnabel2014}.} %FLORS (=FLORS basic).}
\label{tbl:POS-dev-test}
\end{table*}

We observe an asymmetry in the results between some of the domain pairs, e.g. B->D and D->B. We hypothesize that the asymmetry may be due to properties of the data and that the domains are relatively far apart e.g., in terms of $\mathcal{A}$-distance. In fact, asymmetry in these domains is already reflected in the results of \citet{Blitzer2007} and is corroborated in the results for asymmetric tri-training \cite{Saito2017} and our method. 

We note a weakness of this dataset is high variance. Existing approaches only report the mean, which makes an objective comparison difficult. For this reason, we believe it is essential to evaluate proposed approaches also on other tasks.

% ==================== POS tagging results

\paragraph{POS tagging} Results for tagging in the low-data regime (10\% of WSJ) are given in Table \ref{tab:pos-low-data-results}.

Self-training does not work for the sequence prediction task. We report only the best instantiation (throttling with $n$=800). Our results contribute to  negative findings regarding self-training~\cite{Plank2011,VanAsch2016}.

In the low-data setup, tri-training \textit{with disagreement} works best, reaching an overall average accuracy of 89.70, closely followed by classic tri-training, and significantly outperforming the baseline on 4/5 domains. The exception is newsgroups, a difficult domain with high OOV rate where none of the approches beats the baseline (see \S \ref{sec:analysisPos}). Our proposed MT-Tri is better than asymmetric tri-training, but falls below classic tri-training. It beats the baseline significantly on only 2/5 domains (answers and emails). The FLORS tagger~\cite{yin-schnabel-schutze:2015:EMNLP} fares better. Its contextual distributional features are particularly helpful on unknown word-tag combinations (see \S~\ref{sec:analysisPos}), which is a limitation of the lexicalized generic bi-LSTM tagger.

For the high-data setup (Table~\ref{tbl:POS-dev-test}) results are similar. Disagreement, however, is only favorable in the low-data setups; the effect of avoiding easy points no longer holds in the full data setup. Classic tri-training is the best method. In particular, traditional tri-training is complementary to word embedding initialization, pushing the non-pre-trained baseline to the level of \textsc{Src} with Glove initalization. Tri-training pushes performance even further and results in the best model, significantly outperforming the baseline again in 4/5 cases, and reaching FLORS performance on weblogs. Multi-task tri-training is often slightly more effective than asymmetric tri-training~\cite{Saito2017}; however, improvements for both are not robust across domains, sometimes performance even drops. The model likely is too simplistic for such a high-data POS setup, and exploring shared-private models might prove more fruitful~\cite{Liu:ea:2017:ACL}. On the test sets, tri-training performs consistently the best. 

%% POS analysis
\begin{table}%[ht!]
\centering
\resizebox{\columnwidth}{!}{%
\begin{tabular}{l rrrrr}
\toprule
%& \multicolumn{5}{c}{\cellcolor[gray]{.85}\textbf{Target domains}} \\
\textbf{} & \textbf{Ans} & \textbf{Email} & \textbf{Newsg} & \textbf{Rev} & \textbf{Webl}  \\
\% unk tag & 0.25 & 0.80 & 0.31 & 0.06 & 0.0 \\ 
\% OOV  & 8.53 & 10.56 & 10.34 & 6.84 & 8.45 \\
\% UWT & 2.91 & 3.47 & 2.43 & 2.21 & 1.46\\
\midrule
%FLORS* OOV  & 62.15 & 62.61 & 66.42 & 75.29 & 83.64 & 90.37\\
\multicolumn{6}{c}{Accuracy on OOV tokens}\\
Src & 54.26 & 57.48 & \textbf{61.80} & 59.26 & \textbf{80.37} \\
Tri & \textbf{55.53} & \textbf{59.11} & 61.36 & \textbf{61.16} & 79.32\\
Asym   & 52.86 & 56.78 & 56.58 & 59.59 & 76.84\\
MT-Tri & 52.88 & 57.22 & 57.28 & 58.99 & 77.77\\
\midrule
\multicolumn{6}{c}{Accuracy on unknown word-tag (UWT) tokens}\\
Src & \textbf{17.68} & \textbf{11.14} & \textbf{17.88} & \textbf{17.31} & \textbf{24.79} \\
Tri    & 16.88 & 10.04 & 17.58 & 16.35 & 23.65\\
Asym   & 17.16 & 10.43 & 17.84 & 16.92 & 22.74\\
MT-Tri & 16.43 & 11.08 & 17.29 & 16.72 & 23.13\\
\midrule 
FLORS* & 17.19 & 15.13 & 21.97 & 21.06 & 21.65  \\
\bottomrule
\end{tabular}%
}
\caption{Accuracy scores on dev sets for OOV and unknown word-tag (UWT) tokens.} %Above: percentage of unknown tags,\footnote{Unknown tags are: ADD,GW,NFP,XX.} OOV and UWT.}
\label{tbl:POSanalysis}
\end{table}

 \begin{figure}\centering
 \includegraphics[width=1\columnwidth]{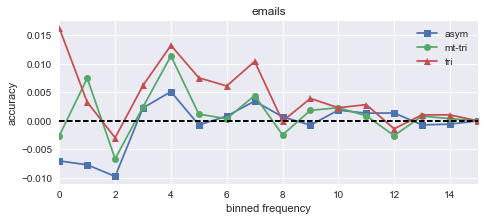}%{logfreq-acc}
 \caption{POS accuracy per binned log frequency.}
 \label{fig:freqacc}
 \end{figure}
 
\paragraph{POS analysis}\label{sec:analysisPos}

We analyze POS tagging accuracy with respect to word frequency\footnote{The binned log frequency was calculated with base 2 (bin 0 are OOVs, bin 1 are singletons and rare words etc).} and unseen word-tag combinations (UWT) on the dev sets. Table~\ref{tbl:POSanalysis} (top rows) provides percentage of unknown tags, OOVs and unknown word-tag (UWT) rate.  The SANCL dataset is overall very challenging: OOV rates are high (6.8-11\% compared to 2.3\% in WSJ), so is the unknown word-tag (UWT) rate (answers and emails contain 2.91\% and 3.47\% UWT compared to 0.61\% on WSJ) and almost all target domains even contain unknown tags~\cite{Schnabel2014} (unknown tags: ADD,GW,NFP,XX), except for weblogs. Email is the domain with the highest OOV rate and highest unknown-tag-for-known-words rate. We  plot accuracy with respect to word frequency on email in Figure~\ref{fig:freqacc}, analyzing how the three methods fare in comparison to the baseline on this difficult domain. 

Regarding OOVs, the results in Table~\ref{tbl:POSanalysis} (second part) show that classic tri-training outperforms the source model (trained on only source data) on 3/5 domains in terms of OOV accuracy, except on two domains with high OOV rate (newsgroups and weblogs).
In general, we note that tri-training works best on OOVs and on low-frequency tokens, which is also shown in  Figure~\ref{fig:freqacc} (leftmost bins). Both other methods fall typically below the baseline in terms of OOV accuracy, but MT-Tri still outperforms Asym in 4/5 cases. Table~\ref{tbl:POSanalysis} (last part) also shows that no bootstrapping method works well on unknown word-tag combinations. UWT tokens are very difficult to predict correctly using an unsupervised approach; the less lexicalized and more context-driven approach taken by FLORS is clearly superior for these cases, resulting in higher UWT accuracies for 4/5 domains.  

% %#B->D,  E->D,  K->D,  D->B,  E->B,  K->B,  B->E,  D->E,  K->E,  B->K,  D->K,  E->K
%     %'SCL':   [78.70, 75.50, 77.00, 78.20, 75.00, 73.00, 75.20, 76.00, 85.20, 77.00, 79.00, 85.00],

% %'BTDNN': [81.90, 78.80, 81.30, 80.40, 77.90, 76.30, 76.90, 78.90, 86.30, 80.10, 82.80, 88.20],

\section{Related work}

\paragraph{Learning under Domain Shift} There is a large body of work on domain adaptation. Studies on unsupervised domain adaptation include early work on \textit{bootstrapping}~\cite{Steedman2003,McClosky2006a}, \textit{shared feature representations}~\cite{Blitzer2006,Blitzer2007} and \textit{instance weighting}~\cite{Jiang2007}. Recent approaches include \textit{adversarial learning}~\cite{Ganin2016} and \textit{fine-tuning}~\cite{sennrich-haddow-birch:2016:P16-11}. There is almost no work on bootstrapping approaches for recent neural NLP, in particular under domain shift. Tri-training is less studied, and only recently re-emerged in the vision community~\cite{Saito2017}, albeit is not compared to classic tri-training.

%There is a large body of work on domain adaptation. Most early work focus on \textit{supervised domain adaptation}~\cite{Daume2007,Jiang2007}, which assumes some labeled target domain data. Studies on \textit{unsupervised domain adaptation}, which we focus on here, include early work on \textit{bootstrapping} approaches~\cite{Steedman2003,McClosky2006a,VanAsch2016,Sogaard2010}, \textit{shared feature representations}~\cite{Blitzer2006,Blitzer2007} or \textit{embeddings} as bridge between domains~\cite{Plank2013} or \textit{fine-tuning}~\cite{sennrich-haddow-birch:2016:P16-11}. There is almost no work examining bootstrapping approaches for recent neural approaches to NLP, which lend itself well for this setup due to their online training nature. In general, the few successful self-training approaches seem to require task-specific components,  like type constraints for POS~\cite{Jorgensen2016} or two-stage parsers~\cite{McClosky2006}. \newcite{VanAsch2016} report an empirically-evaluated success rate of self-training of only 6\% for sentiment analysis. Tri-training approaches are less studied, and only recently re-emerged in the vision community~\cite{Saito2017}. 

\paragraph{Neural network ensembling} Related work on self-ensembling approaches includes snapshot ensembling~\cite{Huang2017c} or temporal ensembling~\cite{Laine2017}. In general, the line between ``explicit'' and ``implicit'' ensembling~\cite{Huang2017c}, like dropout~\cite{Srivastava2014} or temporal ensembling~\cite{Saito2017}, is more fuzzy. As we noted earlier our multi-task learning setup can be seen as a form of self-ensembling.

\paragraph{Multi-task learning in NLP} Neural networks are particularly well-suited for MTL allowing for parameter sharing~\cite{Caruana1993}. Recent NLP conferences witnessed a ``tsunami'' of deep learning papers~\cite{manning2015computational}, followed by what we call a multi-task learning ``wave'':
MTL has been successfully applied to a wide range of NLP tasks~\cite{cohn2013modelling,Hao2015,luong2015multi,Plank2016,fang2016learning,sogaard-goldberg:2016:P16-2,Ruder2017c,Augenstein2018}. 
Related to it is the pioneering work on adversarial learning (DANN)~\cite{Ganin2016}. For sentiment analysis we found tri-training and our MT-Tri model to outperform DANN. Our MT-Tri model lends itself well to shared-private models such as those proposed recently \cite{Liu:ea:2017:ACL,Kim2017b}, which extend upon~\cite{Ganin2016} by having separate source and target-specific encoders.

\section{Conclusions}

We re-evaluate a range of traditional general-purpose bootstrapping algorithms in the context of neural network approaches to semi-supervised learning under domain shift. For the two examined NLP tasks classic tri-training works the best and even outperforms a recent state-of-the-art method. The drawback of tri-training it its time and space complexity. We therefore propose a more efficient multi-task tri-training model, which outperforms both traditional tri-training and recent alternatives in the case of sentiment analysis. For POS tagging, classic tri-training is superior, performing especially well on OOVs and low frequency tokens, which suggests it is less affected by error propagation. Overall we emphasize the importance of comparing neural approaches to strong baselines and reporting results across several runs.

\section*{Acknowledgments}

We thank the anonymous reviewers for their valuable feedback. Sebastian is supported by Irish Research Council Grant Number EBPPG/2014/30 and Science Foundation Ireland Grant Number SFI/12/RC/2289.
Barbara is supported by NVIDIA corporation and thanks the Computing Center of the University of Groningen for HPC support.

\bibliography{semi_supervised}
\bibliographystyle{acl_natbib}

\end{document}